\documentclass{article}
\usepackage{arxiv}

\usepackage[utf8]{inputenc}
\usepackage[T1]{fontenc}
\usepackage{hyperref}
\usepackage{url}
\usepackage{booktabs}
\usepackage{amsfonts}
\usepackage{nicefrac}
\usepackage{microtype}
\usepackage{graphicx}
\usepackage{natbib}
\usepackage{doi}
\usepackage{tikz}
\usetikzlibrary{arrows.meta,positioning,fit,shapes.geometric}
\usepackage{float}
\usepackage{algorithm}
\usepackage{algpseudocode}
\usepackage{amsmath, amssymb, amsthm}
 \usepackage{physics}
\usepackage{algorithm}
\newcommand{\vect}[1]{\mathbf{#1}}
\newcommand{\mat}[1]{\mathbf{#1}}
\usepackage{romannum}
\usepackage{subcaption}
\usepackage{booktabs}
\usepackage{tabularx}
\usepackage{array}

\title{Controllability Analysis of State Space-based Language Model}

\author{
  Mohamed Mabrok \\
  Department of Mathematics\\
  Qatar University\\
  Doha, Qatar \\
  \texttt{m.a.mabrok@qu.edu.qa} \\
  \And
  Yalda Zafari \\
  Department of Mathematics\\
  Qatar University\\
  Doha, Qatar \\
  \texttt{yaldazafari5@gmail.com} \\
}

\begin{document}

\maketitle

\begin{abstract}
\noindent
State-space models (SSMs), particularly Mamba, have emerged as powerful and efficient architectures for sequence modeling. However, their internal dynamics, in contrast to the well-studied attention mechanism, remain opaque. To address this, we validate the \textit{Influence Score}, a novel controllability measure rooted in control-theoretic principles, for SSM-based language models. Derived directly from the Mamba's discretized state-space parameters, the score is computed via a backward recurrence. This recurrence, which is analogous to solving for a system's Observability, quantifies the total influence a token at any position $k$ exerts on all subsequent states and outputs. We deploy this metric in a rigorous comparative study of three different mamba models: \texttt{mamba-130m}, \texttt{mamba-2.8b}, and \texttt{mamba-2.8b-slimpj}. A suite of six experiments analyzes the score's sensitivity to temperature, prompt complexity, token type, layer depth, token position, and input perturbations. Our findings reveal three core insights: \textbf{(1)} The Influence Score scales directly with model size and training data, serving as a robust proxy for model capacity. \textbf{(2)} We identify consistent architectural signatures of Mamba, including a strong recency bias (position sensitivity) and a heavy concentration of influence in mid-to-late layers. \textbf{(3)} We capture emergent behaviors that appear only at scale: the \texttt{mamba-2.8b-slimpj} model is the only one to (a) intuitively rank content words as more influential than function words and (b) demonstrate robustness by decreasing its internal influence when processing noisy inputs. This work validates the Influence Score as a promising  diagnostic tool for interpreting, comparing, and understanding the internal dynamics of SSM-based language models. Our implementation  will be available on  \texttt{\href{https://github.com/Minds-R-Lab/XAI-mamba}{github.com/Minds-R-Lab/XAI-mamba}}
\end{abstract}

\section{Introduction}
The task of modeling sequential data is a cornerstone of modern artificial intelligence, with profound implications across natural language processing, dialogue systems, code generation, and reasoning tasks. For many years, Recurrent Neural Networks (RNNs) and their variants like LSTMs~\cite{hochreiter1997long} were the dominant paradigm for sequence modeling. Their inherent structure, which processes information step-by-step, made them a natural fit for capturing temporal dependencies in language. However, RNNs have well-known limitations, including the vanishing gradient problem, which hampers their ability to learn long-range dependencies, and their sequential nature, which prevents parallelization during training~\cite{bengio1994learning}.

The introduction of the Transformer architecture marked a paradigm shift, largely displacing RNNs in modern large language models (LLMs)~\cite{vaswani2017attention}. By leveraging a self-attention mechanism, Transformers can capture relationships between any two tokens in a sequence, regardless of their distance. This capability to model global context effectively led to unprecedented success in language modeling with GPT~\cite{brown2020language}, BERT~\cite{devlin2019bert}, and their successors, fundamentally transforming how machines process and generate natural language. Despite their groundbreaking performance, Transformers come with a significant drawback: the self-attention mechanism has a computational and memory complexity that scales quadratically ($O(N^2)$) with the sequence length $N$~\cite{tay2022efficient}. This quadratic scaling makes it prohibitively expensive to apply Transformers to long-context scenarios such as book-length document analysis, multi-turn conversations with extensive history, large-scale code repositories, or scientific literature review, precisely the applications where language models promise the greatest impact.

State Space Models (SSMs) have emerged as an alternative approach providing an attractive solution. Drawing from classical control theory originating with Kalman's contributions~\cite{kalman1960contributions}, SSMs process sequences by transforming a one-dimensional input signal (token embeddings) into an output through an intermediate latent state controlled by linear dynamic systems. Initial deep learning implementations of SSMs, like the Structured State Space for Sequences (S4) model, showed these models could function either as recurrent systems for sequential generation or as large convolutional kernels for simultaneous training, allowing them to capture extended dependencies with approximately linear complexity~\cite{gu2022efficiently}. Later iterations, such as S5~\cite{smith2023simplified} and Liquid S4~\cite{hasani2022liquid}, enhanced stability and expressive capability, positioning SSMs as credible alternatives to Transformers in language modeling.

Building upon this foundation, the Mamba architecture introduced a key advancement: a selective scan mechanism~\cite{gu2023mamba}. While earlier SSMs maintained time-invariance (meaning their state transition matrices $\mathbf{A}$, input matrices $\mathbf{B}$, and output matrices $\mathbf{C}$ remained constant), Mamba makes these parameters, particularly the discretization step size $\Delta$, vary based on the input. This capability allows the model to selectively emphasize or disregard particular tokens within the input sequence, efficiently consolidating pertinent context in its hidden state while permitting irrelevant information to fade quickly.

Despite the growing adoption of SSM-based LLMs in production systems, from conversational assistants to code completion tools, a critical gap remains: we lack principled methods to understand what these models have learned and how they process linguistic information. While Transformer-based LLMs benefit from a rich ecosystem of interpretability tools (attention visualization~\cite{clark2019does}, probing classifiers~\cite{belinkov2019analysis}, mechanistic interpretability~\cite{sharkey2025open}), SSMs face challenges due to their fundamentally different information flow. Unlike Transformers, which aggregate information through explicit pairwise attention weights that can be visualized and analyzed, SSMs propagate information \textit{implicitly} through hidden state dynamics governed by differential equations. The question of "which input tokens influenced this generated word?" cannot be answered by inspecting attention matrices because there are none. Instead, we must analyze how input tokens \textit{control} the evolution of the latent state over time, a problem that lies at the intersection of deep learning and dynamical systems theory.

This interpretability gap is particularly acute in high-stakes language applications such as medical report generation, legal document analysis, financial advisory systems, and scientific literature synthesis, where users must understand what the model generated and why it made specific linguistic choices. For instance, if an SSM-based model generates a medical diagnosis from a patient history, clinicians need to know which sentences or phrases in the input most strongly influenced that conclusion and how confident the model was in each piece of evidence. Similarly, in automated code generation, developers must understand which parts of natural language specifications drove specific implementation decisions.

Current explainability approaches, developed mainly for Transformers, fall short when used with SSM-based LLMs due to several core issues. Techniques such as Integrated Gradients~\cite{sundararajan2017axiomatic}, SmoothGrad~\cite{smilkov2017smoothgrad}, and Input $\times$ Gradient~\cite{shrikumar2017learning} calculate token-level importance by assessing how output probabilities respond to input modifications through backpropagation. Yet these techniques are plagued by noisy gradients, saturation problems, and dependence on initial conditions~\cite{adebayo2018sanity}. In SSMs, which incorporate exponential discretization ($\bar{\mathbf{A}} = \exp(\Delta \mathbf{A})$) and complex recurrent state transitions,  gradients may grow unbounded or diminish across extended sequences, generating explanations that fluctuate wildly between runs or under different hyperparameter configurations. Additionally, gradient-based approaches demand numerous forward and backward passes through potentially thousands of tokens, creating substantial computational overhead that undermines the very efficiency benefits that made SSMs attractive initially.

A substantial body of work has focused on interpreting Transformer-based LLMs by analyzing attention patterns~\cite{clark2019does, vig2019visualizing}. Techniques such as attention rollout, attention flow, and attention head analysis provide insights into which tokens the model attends to when making predictions. However, SSMs do not have attention mechanisms, they propagate information through continuous state transitions, not discrete token-to-token interactions. Recent attempts to interpret SSM-based language models have relied on ad-hoc heuristics such as plotting hidden state magnitudes, measuring output sensitivity to token masking~\cite{ali2024hidden}, or probing intermediate representations with linear classifiers~\cite{belinkov2019analysis}. Though these methods offer limited insights, they lack solid theoretical foundations and cannot quantify precisely how much each token governs the model's linguistic reasoning and generation choices.

Even when gradient or perturbation methods yield reasonable token importance rankings, they remain inherently retrospective: they try to deconstruct what the model has already calculated rather than measuring the computation directly.  As Jain and Wallace~\cite{jain2019attention} demonstrated for attention weights in Transformers, high saliency does not guarantee high influence subsequent layers may transform or nullify the contribution of seemingly important features. For SSMs, where linguistic information flows through a cascade of state transitions ($h_t = \bar{\mathbf{A}}h_{t-1} + \bar{\mathbf{B}}u_t$), we require a method that directly measures \textit{controllability}: the extent to which a token at position $t$ can shape the hidden state evolution and thereby influence the model's predictions and generated output at subsequent positions.

In this work, a novel approach for interpreting SSM-based LLMs like Mamba were employed. We utilized the \textbf{Influence Score}  \cite{mabrok2025xvmamba}, a metric derived directly from the Mamba model's discretized state-space parameters ($\bar{\mathbf{A}}$, $\mathbf{B}$, $\mathbf{C}$) at each layer. The score is calculated via a backward recurrence over the token sequence. This computation aggregates the potential impact of an input token $x_k$ on all subsequent states and outputs $\{y_k, y_{k+1}, \dots, y_L\}$. Conceptually, this process is analogous to calculating an observability-weighted measure of controllability for the underlying dynamical system~\cite{kalman1960contributions}. A high Influence Score signifies that a token exerts strong control over the model's future internal state and thereby its generated output. By averaging this score across layers, we obtain a holistic measure of a token's significance within the model's overall computation. This approach offers distinct advantages for understanding SSM-based LLMs:

\begin{enumerate}
    \item The Influence Score is not a post-hoc approximation but is computed directly from the state-space matrices that define the model's dynamics. Its calculation method via backward recurrence is designed to capture the cumulative effect of a token through the system's temporal evolution, providing theoretical grounding for measuring token-level contributions in recurrent state-space architectures.

    \item  By analyzing Influence Scores layer-by-layer, this method can reveal functional specialization within the model's depth. Such analysis has the potential to quantify how information processing is distributed across layers and expose architectural biases, such as positional preferences in how tokens at different locations affect the hidden state trajectory and subsequent predictions.

    \item The Influence Score framework is designed to be sensitive to qualitative differences in model behavior that emerge with scale and training. This enables comparative analysis across model sizes and training regimes, potentially revealing how capabilities like noise handling and semantic understanding develop. The method can be applied across diverse tasks to characterize whether models exhibit task-specific or generalist processing patterns.

    \item  While requiring access to internal activations (captured efficiently via hooks during a standard forward pass), the Influence Score itself is computed post-hoc. The backward recurrence calculation scales linearly with sequence length $L$ and polynomially with the state dimension $N$ (related to matrix operations within the loop). This computational profile is significantly more efficient than perturbation-based methods requiring multiple forward passes, making it practical for analyzing model behavior in real-world applications.
\end{enumerate}

 We validate this framework on three different Mamba-based models: mamba-130m, mamba-2.8b (base), and mamba-2.8b-slimpj.  We conducted comprehensive studies to analyze different characteristics, such as perturbation robustness, token type awareness, layer-wise evolution, positional sensitivity, prompt complexity, and temperature sensitivity using the proposed approach for different Mamba-based models. The paper is organized as follows: Section \ref{sec:background} reviews related work on SSM interpretability and control theory applications. Section \ref{sec:ssm_formulation} presents the controllability formulation and experimental methodology. Section \ref{sec:results_analysis} details results across six experimental dimensions with statistical analysis. Section \ref{sec:disscution} discusses implications and applications. Section \ref{sec:disscution} concludes with future directions.

\section{Background and Related Work}
\label{sec:background}
\subsection{State Space Models: From Signal Processing to Language}

For years, sequence modeling has been dominated by the Transformer architecture \cite{vaswani2017attention}, which relies on a self-attention mechanism. While powerful, self-attention's $O(L^2)$ computational and memory complexity with respect to sequence length $L$ creates a significant bottleneck for processing very long contexts. State Space Models have recently emerged as a highly efficient alternative, drawing inspiration from classical models in signal processing and control theory. A continuous-time SSM is defined by a linear ordinary differential equation (ODE):
\begin{equation}
\begin{aligned}
h'(t) &= \mathbf{A}h(t) + \mathbf{B}x(t) \\
y(t) &= \mathbf{C}h(t) + \mathbf{D}x(t)
\end{aligned}
\label{eq:ssm_continuous}
\end{equation}
where $x(t) \in \mathbb{R}^{M}$ is the input, $h(t) \in \mathbb{R}^{N}$ is the latent state, $y(t) \in \mathbb{R}^{M}$ is the output, and $\mathbf{A}, \mathbf{B}, \mathbf{C}, \mathbf{D}$ are the state-space matrices. To be used in deep learning, this continuous system must be discretized using a timestep $\Delta$, which transforms it into a discrete-time recurrence:
\begin{equation}
\begin{aligned}
h_k &= \bar{\mathbf{A}}h_{k-1} + \bar{\mathbf{B}}x_k \\
y_k &= \bar{\mathbf{C}}h_k + \bar{\mathbf{D}}x_k
\end{aligned}
\label{eq:ssm_discrete}
\end{equation}
where $(\bar{\mathbf{A}}, \bar{\mathbf{B}}, \bar{\mathbf{C}}, \bar{\mathbf{D}})$ are the discretized matrices derived from $(\mathbf{A}, \mathbf{B}, \mathbf{C}, \mathbf{D}, \Delta)$. This recurrent form (Eq. \ref{eq:ssm_discrete}) is computationally efficient for inference, scaling linearly $O(L)$ with sequence length. A key innovation of modern SSMs, such as S4 \cite{gu2021efficiently}, is that this system can also be represented as a discrete convolution, enabling a highly parallel $O(L \log L)$ training mode. The Mamba architecture \cite{gu2023mamba} overcomes a critical limitation of prior SSMs: their time-invariant (LTI) nature. In Mamba, the state-space matrices $\mathbf{A}$, $\mathbf{B}$, $\mathbf{C}$ and the timestep $\Delta$ are not fixed but are instead \textit{functions of the input $x_k$}. This \textit{selectivity} allows the model to dynamically adjust its state dynamics based on the current token, enabling it to "select" or "forget" information from its hidden state---a capability that is crucial for complex language tasks.



\subsection{ Interpretability Methods}
Most existing work on LLM interpretability focuses on the Transformer:
\begin{itemize}
    \item {Attention Analysis:} Visualizing attention weights \cite{vaswani2017attention} has been a common (though debated) method for understanding information flow. This is fundamentally \textit{inapplicable} to SSMs, which have no such mechanism.
    \item {Gradient-based Methods:} Techniques like Saliency and Integrated Gradients \cite{sundararajan2017axiomatic} are model-agnostic and can be applied to Mamba, but they only provide a post-hoc sensitivity analysis rather than probing the model's internal state mechanism.
    \item {Probing Tasks:} Training simple linear classifiers on intermediate representations \cite{belinkov2019analysis} can reveal what information is encoded in the hidden state $h_k$, but not \textit{how} that state is controlled or evolved.
    \item {Mechanistic Interpretability:} Circuit-level analysis \cite{sharkey2025open} aims to reverse-engineer specific sub-graphs within a model, a methodology that is complex and has not yet been systematically applied to SSM architectures.
\end{itemize}

\subsection{ Control Theoretic Background}
 An SSM (Eq. \ref{eq:ssm_continuous}) is a \textit{dynamical system}, and its properties have been studied for over a century in control theory. In particular, we capitalize on two core concepts controllability and observability.

Controllability addresses the question: \textit{Can we steer the internal state $h(t)$ of the system from any initial state $h(t_0)$ to any desired final state $h(t_f)$ within a finite time interval $[t_0, t_f]$, using some admissible control input $x(t)$?} In simpler terms, it measures the degree to which the input $x(t)$ can influence or control the evolution of the internal state $h(t)$.

For Linear Time-Invariant (LTI) systems (where $\mathbf{A}$, $\mathbf{B}$, $\mathbf{C}$, $\mathbf{D}$ are constant), a system is completely controllable if and only if the controllability matrix $\mathcal{C}$ has full rank (rank $N$, where $N$ is the dimension of the state vector $h(t)$):
\begin{equation}
\label{eq:controllability_matrix}
\mathcal{C} = \begin{bmatrix} \mathbf{B} & \mathbf{A}\mathbf{B} & \mathbf{A}^2\mathbf{B} & \cdots & \mathbf{A}^{N-1}\mathbf{B} \end{bmatrix}
\end{equation}
The rank condition ensures that the inputs, propagated through the system dynamics ($\mathbf{A}$), can span the entire state space.

A related and often more useful concept, especially for continuous-time systems over an interval $[t_0, t_f]$, is the Controllability Gramian, denoted $\mathcal{W}_c(t_0, t_f)$:
\begin{equation}
\label{eq:controllability_gramian}
\mathcal{W}_c(t_0, t_f) = \int_{t_0}^{t_f} e^{\mathbf{A}(t_0 - \tau)} \mathbf{B} \mathbf{B}^T e^{\mathbf{A}^T(t_0 - \tau)} d\tau
\end{equation}
The system is controllable on $[t_0, t_f]$ if and only if $\mathcal{W}_c(t_0, t_f)$ is invertible (positive definite). The magnitude of the eigenvalues of $\mathcal{W}_c$ quantifies the energy required to steer the state in different directions; directions corresponding to small eigenvalues are hard to control. In the context of Mamba, high controllability implies that the input tokens $x(t)$ have a strong capacity to shape the internal hidden state $h(t)$.

Observability addresses the dual question: \textit{Can we determine the initial internal state $h(t_0)$ of the system by observing only the system's output $y(t)$ and input $x(t)$ over a finite time interval $[t_0, t_f]$?} It measures the degree to which the internal state $h(t)$ is reflected in or can be inferred from the output $y(t)$.

For LTI systems, a system is completely observable if and only if the observability matrix $\mathcal{O}$ has full rank (rank $N$):
\begin{equation}
\label{eq:observability_matrix}
\mathcal{O} = \begin{bmatrix} \mathbf{C} \\ \mathbf{C}\mathbf{A} \\ \mathbf{C}\mathbf{A}^2 \\ \vdots \\ \mathbf{C}\mathbf{A}^{N-1} \end{bmatrix}
\end{equation}
This condition ensures that the internal state dynamics, projected onto the output via $\mathbf{C}$, are sufficiently distinct to allow reconstruction of the initial state.

Similar to controllability, the Observability Gramian, denoted $\mathcal{W}_o(t_0, t_f)$, provides insights for continuous-time systems:
\begin{equation}
\label{eq:observability_gramian}
\mathcal{W}_o(t_0, t_f) = \int_{t_0}^{t_f} e^{\mathbf{A}^T(\tau - t_0)} \mathbf{C}^T \mathbf{C} e^{\mathbf{A}(\tau - t_0)} d\tau
\end{equation}
The system is observable on $[t_0, t_f]$ if and only if $\mathcal{W}_o(t_0, t_f)$ is invertible (positive definite). Small eigenvalues correspond to state components that are hard to observe from the output. In Mamba, high observability implies that the internal state $h(t)$ strongly manifests in the output logits $y(t)$, making the internal processing visible in the predictions. While state controllability focuses on reaching any state $h(t_f)$, output controllability is concerned with reaching any desired output $y(t_f)$. A system is output controllable if it's possible to steer the output $y(t)$ from any initial output $y(t_0)$ to any final output $y(t_f)$ in finite time using some input $x(t)$.

For systems like Mamba where $\mathbf{D}$ is typically zero or negligible in the core SSM layer (it's often handled by residual connections), output controllability is closely tied to state controllability. If we can control the state $h(t)$ via $\mathbf{B}$ and the state strongly influences the output via $\mathbf{C}$ (i.e., the system is observable), then we can effectively control the output $y(t)$. These concepts are crucial for interpreting Mamba's dynamics. The Influence Score introduced in this work can be understood as a practical, computationally tractable measure that approximates an observability-weighted controllability. The backward recurrence calculation inherently combines:
\begin{enumerate}
    \item The ability of an input $x_k$ to affect the state $h_k$ (related to $\mathbf{B}_k$).
    \item The persistence of that effect through future states $h_{j>k}$ (related to $\mathbf{A}_j$).
    \item The degree to which those future states $h_j$ manifest in the outputs $y_j$ (related to $\mathbf{C}_j$).
\end{enumerate}
By analyzing this score, we gain insights into which inputs exert significant control over the model's internal processing trajectory and its final output, moving beyond generic methods to leverage the specific mathematical structure of SSMs. While these concepts have been used to analyze the stability and dynamics of simple RNNs, their application to modern, large-scale, selective SSMs remains a nascent field. This presents a significant gap: existing interpretability tools are either inapplicable (attention) or fail to exploit the model's underlying mathematical structure (gradients, probes).

A recent study \cite{mabrok2025xvmamba} filled this gap by introducing the Influence Score, a novel metric derived directly from control-theoretic principles. By computing a backward recurrence analogous to solving for a system's Observability Gramian, the score quantifies the influence a given token $x_k$ has on all future system outputs $y_{j>k}$. It is a principled measure of token-level controllability and observability within the Mamba architecture, providing a new lens that is uniquely specialized to SSM-based models.

\section{Methodology}
\label{sec:ssm_formulation}
The fundamental flaw in measuring influence on the final hidden state, $\vec{x}_L$, is that it is not the direct input to the classifier. Vision State Space Models (SSMs) almost universally employ a Global Average Pooling (GAP) layer after the final SSM layer. The GAP layer computes the mean of the output vectors, $\vec{y}_k$, across all $L$ sequence steps (patches). This produces a single, fixed-size feature vector, $\bar{\vec{y}}$, which is then fed into the final linear classification head.
\begin{equation}
    \bar{\vec{y}} = \frac{1}{L} \sum_{j=1}^{L} \vec{y}_j = \frac{1}{L} \sum_{j=1}^{L} (\bar{\mathbf{C}}_j \vec{x}_j + \bar{\mathbf{D}}_j \vec{u}_j)
    \label{eq:gap}
\end{equation}
Therefore, to create a meaningful interpretability metric, we must measure the influence of an input patch $\vec{u}_k$ on this aggregated vector $\bar{\vec{y}}$. An input $\vec{u}_k$ influences the aggregated output $\bar{\vec{y}}$ by contributing to its own output term, $\vec{y}_k$, and to all subsequent output terms, $\vec{y}_j$ where $j > k$. The total influence is the sum of the magnitudes of these individual contributions. The state $\vec{x}_j$ at any step $j$ is a function of a prior state $\vec{x}_k$ ($k < j$) and the intervening inputs:
\begin{equation}
    \vec{x}_j = \left( \prod_{i=k}^{j-1} \bar{\mathbf{A}}_i \right) \vec{x}_k + \sum_{m=k}^{j-1} \left( \prod_{i=m+1}^{j-1} \bar{\mathbf{A}}_i \right) \bar{\mathbf{B}}_m \vec{u}_m
\end{equation}
The partial derivative $\frac{\partial \vec{x}_j}{\partial \vec{u}_k}$ gives us the linear transformation that maps a change in input $\vec{u}_k$ to a change in a future state $\vec{x}_j$:
\begin{equation}
    \frac{\partial \vec{x}_j}{\partial \vec{u}_k} = \left( \prod_{i=k+1}^{j-1} \bar{\mathbf{A}}_i \right) \bar{\mathbf{B}}_k \quad \text{for } j > k
    \label{eq:state_jacobian}
\end{equation}
Similarly, the influence of $\vec{u}_k$ on an output $\vec{y}_j$ is found by applying the chain rule:
\begin{equation}
    \frac{\partial \vec{y}_j}{\partial \vec{u}_k} = \frac{\partial \vec{y}_j}{\partial \vec{x}_j} \frac{\partial \vec{x}_j}{\partial \vec{u}_k} = \bar{\mathbf{C}}_j \left[ \left( \prod_{i=k+1}^{j-1} \bar{\mathbf{A}}_i \right) \bar{\mathbf{B}}_k \right]
    \label{eq:output_jacobian}
\end{equation}
For the special case where $j=k$, the influence is direct (it does not pass through $\mathbf{A}$):
\begin{equation}
    \frac{\partial \vec{y}_k}{\partial \vec{u}_k} = \bar{\mathbf{C}}_k \frac{\partial \vec{x}_k}{\partial \vec{u}_k} + \bar{\mathbf{D}}_k = \bar{\mathbf{C}}_k \bar{\mathbf{B}}_k + \bar{\mathbf{D}}_k
\end{equation}
Assuming the feedthrough matrix $\mathbf{D}$ is negligible or zero (common in these architectures), this simplifies to $\bar{\mathbf{C}}_k \bar{\mathbf{B}}_k$.

The total influence of $\vec{u}_k$ is the sum of the norms (we use the Frobenius norm, $\|\cdot\|_F$) of these Jacobian matrices for all outputs from $k$ to $L$:
\begin{equation}
    \text{InfluenceScore}(k) = \sum_{j=k}^{L} \left\| \frac{\partial \vec{y}_j}{\partial \vec{u}_k} \right\|_F
\end{equation}
Expanding this gives our final, detailed formula:
\begin{equation}
    \text{InfluenceScore}(k) = \underbrace{\| \bar{\mathbf{C}}_k \bar{\mathbf{B}}_k \|_F}_{\text{Term 1: Direct Influence}} + \underbrace{\sum_{j=k+1}^{L} \left\| \bar{\mathbf{C}}_j \left( \prod_{i=k+1}^{j-1} \bar{\mathbf{A}}_i \right) \bar{\mathbf{B}}_k \right\|_F}_{\text{Term 2: Propagated Influence}}
    \label{eq:final_detailed}
\end{equation}

\paragraph{Term 1: The Direct Influence $\|\bar{\mathbf{C}}_k \bar{\mathbf{B}}_k\|_F$}
This term measures the immediate impact of the input $\vec{u}_k$ on its own corresponding output $\vec{y}_k$.
\begin{itemize}
    \item $\bar{\mathbf{B}}_k$: This matrix transforms the input patch $\vec{u}_k$ into a change in the hidden state $\vec{x}_k$. A large norm for $\bar{\mathbf{B}}_k$ means the input is salient and can strongly perturb the state.
    \item $\bar{\mathbf{C}}_k$: This matrix reads out the hidden state $\vec{x}_k$ to produce the output $\vec{y}_k$. A large norm for $\bar{\mathbf{C}}_k$ means the current state is considered important for the output.
\end{itemize}
The product $\bar{\mathbf{C}}_k \bar{\mathbf{B}}_k$ represents the full, instantaneous input-output path at step $k$. This term ensures that every patch has a baseline influence score, preventing the "vanishing influence" problem for early patches.

\paragraph{Term 2: The Propagated Influence}
This term is a sum that captures the long-term impact of $\vec{u}_k$ on all subsequent outputs. For each future step $j > k$, the matrix product within the norm represents the dynamical path from input $\vec{u}_k$ to output $\vec{y}_j$.
\begin{itemize}
    \item $\left( \prod_{i=k+1}^{j-1} \bar{\mathbf{A}}_i \right)$: This is the state transition matrix that describes how the initial state perturbation caused by $\vec{u}_k$ evolves and propagates through the system's memory over time.
    \item The entire term $\bar{\mathbf{C}}_j (\dots) \bar{\mathbf{B}}_k$ is the linear operator mapping the input at time $k$ to the output at a future time $j$.
\end{itemize}
This sum  solves the recency bias. An early patch ($k \ll L$) has its influence on any single distant output $\vec{y}_L$ diminished by decay, but it has the opportunity to influence a large number of subsequent outputs, and the sum of these contributions can be substantial.

\subsection{Influence Score Algorithm for Language SSMs}
The following algorithm details the computation of the token-level Influence Score used in this pipeline, derived from the Mamba layer's state-space parameters.

\begin{algorithm}[H]
\caption{Token Influence Score Calculation for Mamba Layers}
\label{alg:mamba_influence_score}
\begin{algorithmic}[1]
\State \textbf{Input:} Hidden state sequence $\vect{X} \in \mathbb{R}^{B \times L \times D}$ (Batch Size $B$, Seq Len $L$, Model Dim $D$) for a single Mamba layer.
\State \textbf{Output:} Influence Score vector $\vect{S} \in \mathbb{R}^{L}$ (One score per token, averaged over batch and state dimensions).

\State \Comment{Project input and compute SSM parameters}
\State $\vect{X}_{in}, \vect{Z} \leftarrow \text{Split}(\text{LinearLayer}_{in}(\vect{X}))$
\State $\vect{X}_{conv} \leftarrow \text{Conv1D}(\vect{X}_{in})$
\State $\vect{X}_{act} \leftarrow \text{Activation}(\vect{X}_{conv})$
\State $\Delta_k^{raw}, \vect{B}_k^{raw}, \vect{C}_k^{raw} \leftarrow \text{Split}(\text{LinearLayer}_{proj}(\vect{X}_{act}))$ \Comment{Indices $k=1 \dots L$}
\State $\Delta_k \leftarrow \text{Softplus}(\text{LinearLayer}_{\Delta}(\Delta_k^{raw}))$
\State $\vect{A} \leftarrow -\exp(\vect{A}_{log})$ \Comment{Shared A matrix from layer parameters}

\State \Comment{Discretize A and take absolute values}
\State $\bar{\mat{A}}_k \leftarrow \exp(\Delta_k \otimes \mat{A})$ \Comment{Outer product for each token $k$}
\State $\mat{A}_{abs} \leftarrow |\bar{\mat{A}}|$ \Comment{Shape: $B \times L \times N \times N$, $N$=state dim}
\State $\mat{B}_{abs} \leftarrow |\text{Unsqueeze}(\vect{B}^{raw})|$ \Comment{Shape: $B \times L \times N \times 1$}
\State $\mat{C}_{abs} \leftarrow |\text{Unsqueeze}(\vect{C}^{raw})|$ \Comment{Shape: $B \times L \times N \times 1$}

\State \Comment{Backward recurrence to compute influence}
\State Initialize score tensor $\mat{I} = \vect{0}^{B \times L \times N \times 1}$
\State Initialize future influence propagator $\mat{P} = \vect{0}^{B \times N \times 1}$
\For{$k = L-2 \to 0$ (iterating backwards)}
    \State $\mat{P} \leftarrow \mat{C}_{abs}[:, k+1, :, :] + \mat{A}_{abs}[:, k+1, :, :] \odot \mat{P}$ \Comment{$\odot$ is element-wise or matmul depending on P shape}
    \State $\text{Term\_1}_k \leftarrow |\mat{C}_{abs}[:, k, :, :] \odot \mat{B}_{abs}[:, k, :, :]|$ \Comment{Direct influence, element-wise}
    \State $\text{Term\_2}_k \leftarrow \mat{B}_{abs}[:, k, :, :] \odot \mat{P}$ \Comment{Propagated influence, element-wise}
    \State $\mat{I}[:, k, :, :] \leftarrow \text{Term\_1}_k + \text{Term\_2}_k$
\EndFor
\State $\text{Term\_1}_{L-1} \leftarrow |\mat{C}_{abs}[:, L-1, :, :] \odot \mat{B}_{abs}[:, L-1, :, :]|$ \Comment{Handle last token}
\State $\mat{I}[:, L-1, :, :] \leftarrow \text{Term\_1}_{L-1}$

\State \Comment{Aggregate scores}
\State $\vect{S}_{agg} \leftarrow \sum_{\text{dim}=N} \mat{I}$ \Comment{Sum over state dimension N, result shape $B \times L \times 1$}
\State $\vect{S}_{final} \leftarrow \text{Mean}_{\text{dim}=B, \text{dim}=1} (\vect{S}_{agg})$ \Comment{Average over batch and remaining dim, result shape $L$}
\State $\vect{S} \leftarrow \vect{S}_{final}$
\State \textbf{return} $\vect{S}$
\end{algorithmic}
\end{algorithm}

Our experimental design is structured to comprehensively evaluate the behavior, robustness, and internal dynamics of Mamba-based models using the proposed Jacobian-based Influence Score. The design consists of three core components: the models under evaluation, the metric implementation, and a diverse suite of six targeted experiments.

\subsection{Models and Metric Implementation}
We evaluate three Mamba-based models, selected to represent variations in both scale and training data quality:
\begin{itemize}
    \item \textbf{M1: `mamba-2.8b-slimpj`}: A 2.8B parameter model trained on the high-quality, filtered SlimPajama dataset.
    \item \textbf{M2: `mamba-2.8b`}: A 2.8B parameter model trained on a base, general-purpose dataset.
    \item \textbf{M3: `mamba-130m`}: A 130M parameter model, serving as a baseline for scale comparison.
\end{itemize}

Our core metric, the \textit{Influence Score} (defined in Section\ref{sec:ssm_formulation}), is computed by dynamically patching the models at runtime. We implement this by wrapping each Mamba layer (specifically, the `mamba\_ssm.modules.mamba\_simple.Mamba` mixer) in a custom `ControllabilityMamba` module. During the forward pass, we use PyTorch hooks via the `capture\_and\_analyze` function to intercept the input tensor $x$ for each layer. This tensor is then processed by the `compute\_controllability` method, which calculates the influence score for the entire sequence using an efficient $O(L)$ reverse-scan algorithm based on the derived state-space parameters ($\bar{\mathbf{A}}$, $\bar{\mathbf{B}}$, $\bar{\mathbf{C}}$). The final score for each experiment is the mean influence calculated across all tokens in the generated sequence (prompt + new tokens).

\subsection{Experimental Suite}

\noindent
The \textit{Influence Score} quantifies the contribution of each token and layer to the model’s generative dynamics. It is computed via feature perturbation and gradient attribution on the model’s hidden state transitions, producing per-token and per-layer influence maps. Averaging these over runs yields stable influence distributions. The Table in \ref{tab:experimental_suite_updated} summarizes the six experiments conducted to evaluate the model's behavior and the robustness of the Influence Score metric. The central goal is to evaluate whether the Influence Score captures interpretable patterns across layers, token types, prompts, and model scales, and whether it can serve as a complementary diagnostic tool to existing attention-based interpretability analyses.

\begin{figure}[ht]
    \centering
    \includegraphics[width=0.5\linewidth]{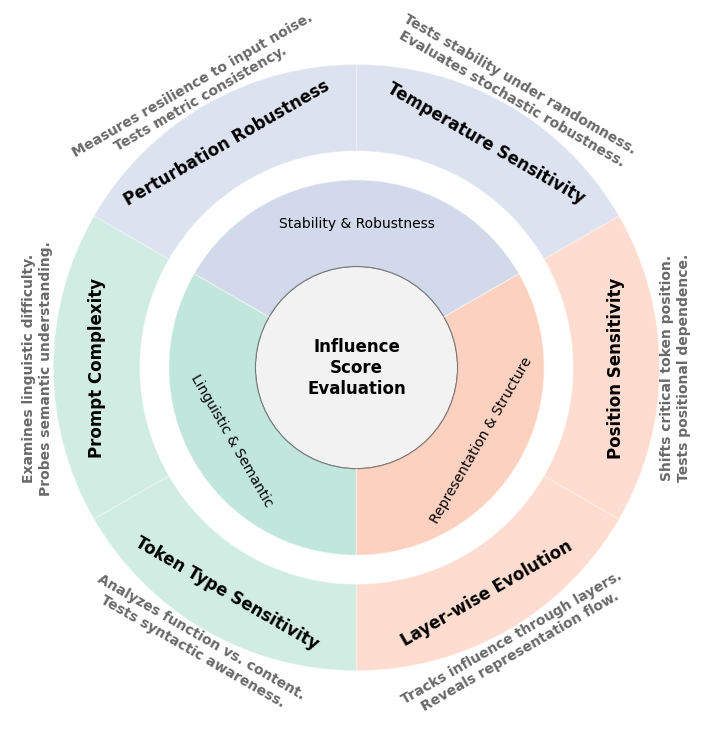}
    \caption{Suite of Experiments for Influence Score Evaluation. For each experimental condition, we report the  mean and standard deviation of the Influence Score, the coefficient of variation (CV) as a stability indicator, and the spearman correlation coefficients for monotonic trends (e.g., temperature vs. influence).}
    \label{fig:placeholder}
\end{figure}

\begin{table}[!ht]
\centering
\small
\begin{tabularx}{\textwidth}{p{0.5cm} p{2.5cm} p{4cm} X}
\toprule
\textbf{Exp.} & \textbf{Objective} & \textbf{Key Parameters} & \textbf{Prompt Examples / Variations (N=10 per category)} \\
\midrule

1 & \textbf{Temperature Sensitivity:} Evaluate the stability of the Influence Score metric under varying sampling stochasticity.
& \texttt{top\_p=0.9}, \texttt{rep\_penalty=1.1}, \texttt{max\_new\_tokens=30}, varying \texttt{temperature} $\in \{0.3, 0.5, 0.7, 1.0, 1.5\}$
& 
``The capital of France is'' \\

\midrule

2 & \textbf{Prompt Complexity:} Quantify the impact of linguistic and cognitive complexity on influence magnitude and variability.
& \texttt{temp=0.7}, \texttt{top\_p=0.9}, \texttt{rep\_penalty=1.2}, \texttt{max\_new\_tokens=40}
& \textbf{Factual:} ``The chemical formula for water is'' \newline ``What is the largest planet in our solar system?'' \newline
\textbf{Reasoning:} ``If all humans are mortal, and Socrates is human, then'' \newline ``A is taller than B, and B is taller than C. Therefore,'' \newline
\textbf{Creative:} ``Once upon a time in a magical forest,'' \newline ``The spaceship landed on the alien planet...'' \newline
\textbf{Technical:} ``To implement a binary search tree in Python, first'' \newline ``A major difference between TCP and UDP is that'' \newline
\textbf{Ambiguous:} ``The bank is'' \newline ``She saw the man with the'' \\

\midrule

3 & \textbf{Token Type Sensitivity:} Examine the distribution of influence across syntactic categories (function vs. content).
& \texttt{temp=0.7}, \texttt{top\_p=0.9}, \texttt{rep\_penalty=1.0}, \texttt{max\_new\_tokens=40}
& \textbf{Content-heavy:} ``Artificial intelligence research focuses on developing'' \newline ``Quantum mechanics describes the physical properties of'' \newline
\textbf{Function-heavy:} ``The of the and for the in the to the'' \newline ``It is a fact that there is no one who'' \newline
\textbf{Mixed:} ``She walked into the room and saw something extraordinary'' \newline ``The old sailor looked at the storm and sighed'' \\

\midrule

4 & \textbf{Layer-wise Evolution:} Trace the progression of token influence through early, middle, and late model layers.
& \texttt{temp=0.7}, \texttt{top\_p=0.9}, \texttt{rep\_penalty=1.2}, \texttt{max\_new\_tokens=30}
& \textbf{Simple:} ``The cat sat on the'' \newline ``One plus one equals'' \newline
\textbf{Complex:} ``The philosophical implications of artificial intelligence include'' \newline ``The geopolitical ramifications of the energy crisis are'' \newline
\textbf{Technical:} ``To optimize neural network training, we should'' \newline ``The primary function of a CPU is to'' \\

\midrule

5 & \textbf{Position Sensitivity:} Assess the impact of critical information position (front, back, or distributed) on token influence.
& \texttt{temp=0.7}, \texttt{top\_p=0.9}, \texttt{rep\_penalty=1.2}, \texttt{max\_new\_tokens=30}
& \textbf{Front-critical:} ``\textbf{INSTRUCTION: Translate...} 'Hello, how are you?'" \newline ``\textbf{CRITICAL: The password is 'Omega'}. All other...'' \newline
\textbf{Back-critical:} ``The following text is a simple greeting... \textbf{INSTRUCTION: 'Hello, how are you?'}" \newline ``All other information is irrelevant... \textbf{CRITICAL: This is the password.}'' \newline
\textbf{Distributed:} ``\textbf{INSTRUCTION:} Translate 'Hello, how are you?' to French. \textbf{TASK: Translation.}'' \\

\midrule

6 & \textbf{Perturbation Robustness:} Measure resilience of Influence Scores under input-level modifications (linguistic noise).
& \texttt{temp=0.7}, \texttt{top\_p=0.9}, \texttt{rep\_penalty=1.2}, \texttt{max\_new\_tokens=20}
& \textit{Example Set 1 (N=10 total):} \newline
\textbf{Original:} ``The first man on the moon was Neil Armstrong'' \newline
\textbf{Remove article:} ``First man on moon was Neil Armstrong'' \newline
\textbf{Typo:} ``The first man \textbf{om} the moon \textbf{iz} Neil Armstrong'' \newline
\textbf{Synonym:} ``The first \textbf{person} on the moon was Neil Armstrong'' \newline
\textbf{Reorder:} ``Neil Armstrong moon was on the The first man'' \\

\bottomrule
\end{tabularx}
\caption{Summary of the six experiments designed to evaluate the \textbf{Influence Score} across stochastic, linguistic, structural, and robustness dimensions in SSM-based LLMs. Each experiment isolates a specific representational factor while maintaining consistent generation settings.}
\label{tab:experimental_suite_updated}
\end{table}

\section{Results and Analysis}
\label{sec:results_analysis}
This section presents the empirical findings from applying the Influence Score metric across our suite of six experiments. We analyze the results obtained from three Mamba models, \texttt{mamba-130m}, \texttt{mamba-2.8b}, and \texttt{mamba-2.8b-slimpj}, evaluating their performance and internal dynamics under varying conditions of temperature, prompt complexity, token type distribution, layer depth, positional information, and input perturbations. Our analysis focuses on comparing the models to reveal how the Influence Score captures scaling effects, consistent architectural signatures, and emergent behaviors related to training data quality and model robustness.
\subsection{Overview}
\begin{figure}[!ht]
    \centering
    \includegraphics[width=0.9\linewidth]{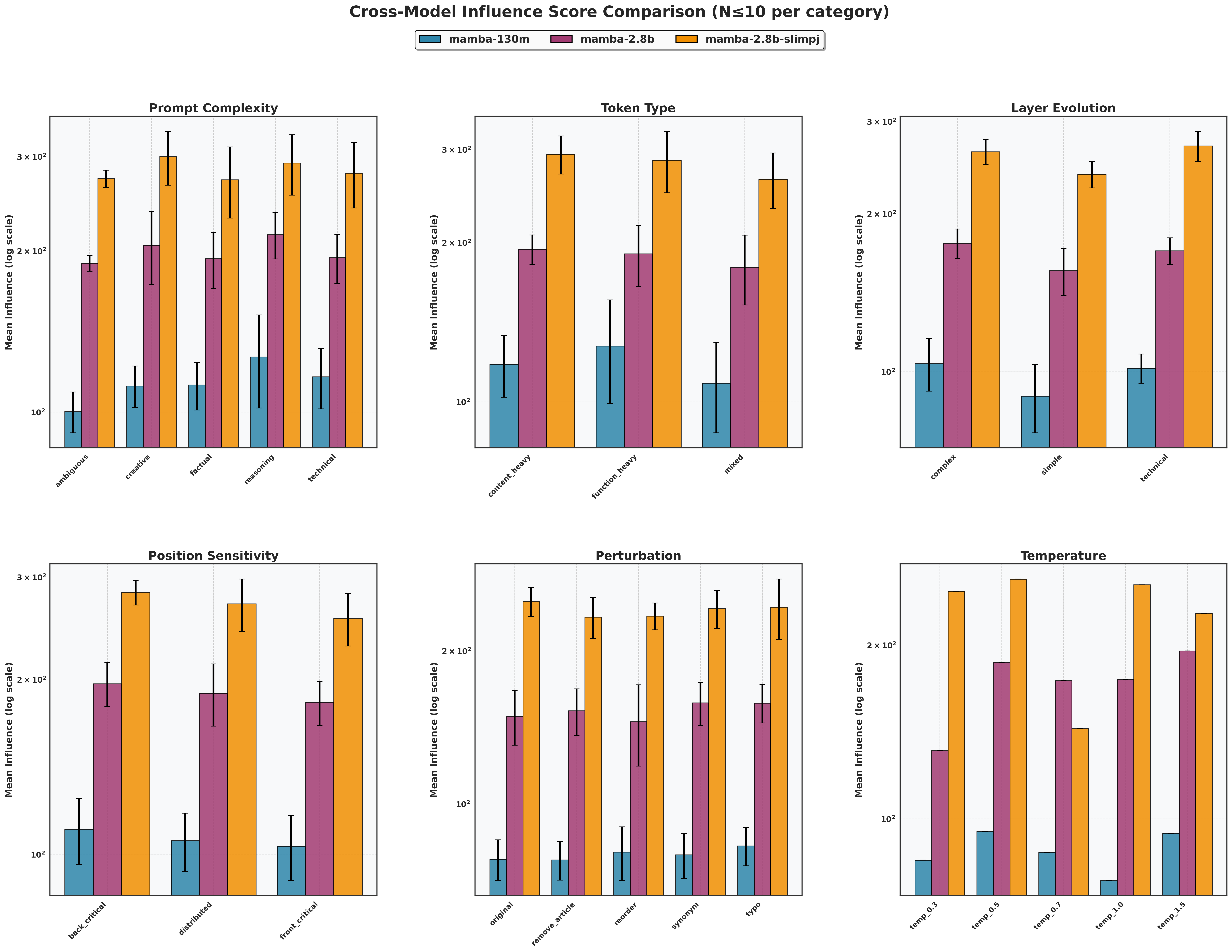}
    \caption{Cross-model comparison of mean Influence Scores across all six experiment categories.}
    \label{fig:cross_model_comparison}
\end{figure}

Figure~\ref{fig:cross_model_comparison} provides an overview of cross-model trends across the six experiments. 
A consistent scaling effect is observed: the mean Influence Score increases monotonically with model size, confirming that larger Mamba variants maintain stronger internal token dependencies and more stable representational hierarchies.

\subsection{Temperature Sensitivity}
\label{subsec:temperature}
Across all models, the Influence Score remained relatively stable across temperatures (Spearman $\rho = 0.0$ for Mamba-130M, $\rho = 0.7$ for Mamba-2.8B, and $\rho = -0.3$ for Mamba-2.8B-SlimPJ). This indicates that the metric is largely insensitive to sampling stochasticity. However, larger models exhibit mild positive correlation with temperature, suggesting enhanced adaptability in higher-entropy sampling regimes.

\begin{figure}[!ht]
    \centering
    \includegraphics[width=0.9\linewidth]{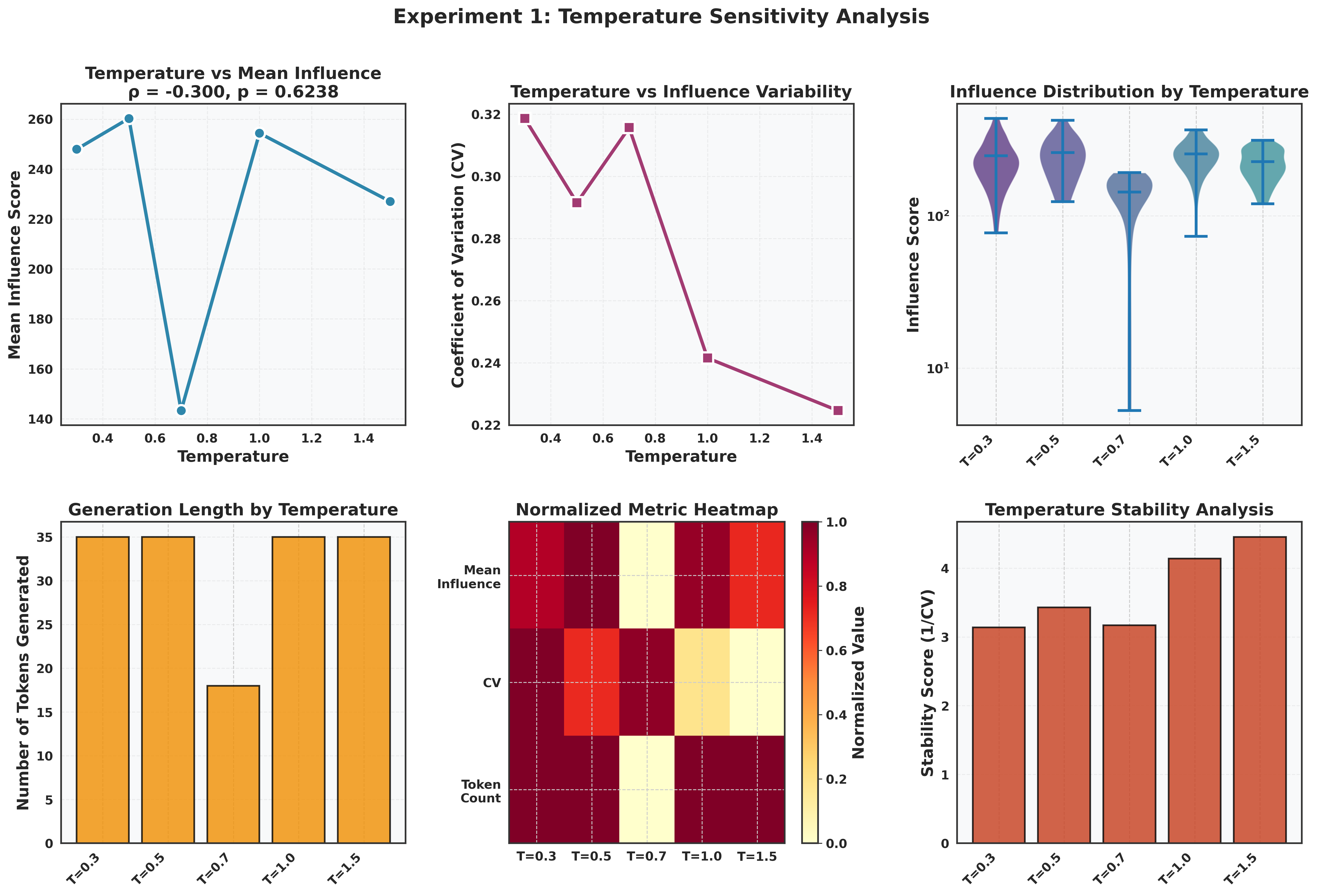}
    \caption{Temperature sensitivity for the Mamba-2.8B-SlimPJ model. Each line represents mean Influence Score per temperature setting.}
    \label{fig:temperature_analysis}
\end{figure}

\subsection{Prompt Complexity}
\label{subsec:prompt_complexity}
Influence increased systematically with prompt reasoning complexity across all scales (Table~\ref{tab:prompt_complexity}). 
Creative and reasoning prompts consistently triggered the highest influence values, reflecting deeper representational dynamics and higher inter-token coupling in abstract reasoning contexts.
The 2.8B-SlimPJ model achieved the highest mean influence ($3.00\times10^2$), highlighting its capacity for structured thought and long-range token dependencies.

\begin{table}[!ht]
\centering
\caption{Mean Influence Score per prompt type. Values are averaged over 10 runs per category.}
\label{tab:prompt_complexity}
\begin{tabular}{lccc}
\toprule
\textbf{Prompt Type} & \textbf{Mamba-130M} & \textbf{Mamba-2.8B} & \textbf{Mamba-2.8B-SlimPJ} \\
\midrule
Reasoning  & 1.27e+02 & 2.14e+02 & 2.92e+02 \\
Creative   & 1.12e+02 & 2.05e+02 & 3.00e+02 \\
Technical  & 1.16e+02 & 1.94e+02 & 2.79e+02 \\
Factual    & 1.12e+02 & 1.93e+02 & 2.71e+02 \\
Ambiguous  & 1.00e+02 & 1.90e+02 & 2.73e+02 \\
\bottomrule
\end{tabular}
\end{table}

\begin{figure}[!ht]
    \centering
    \includegraphics[width=0.9\linewidth]{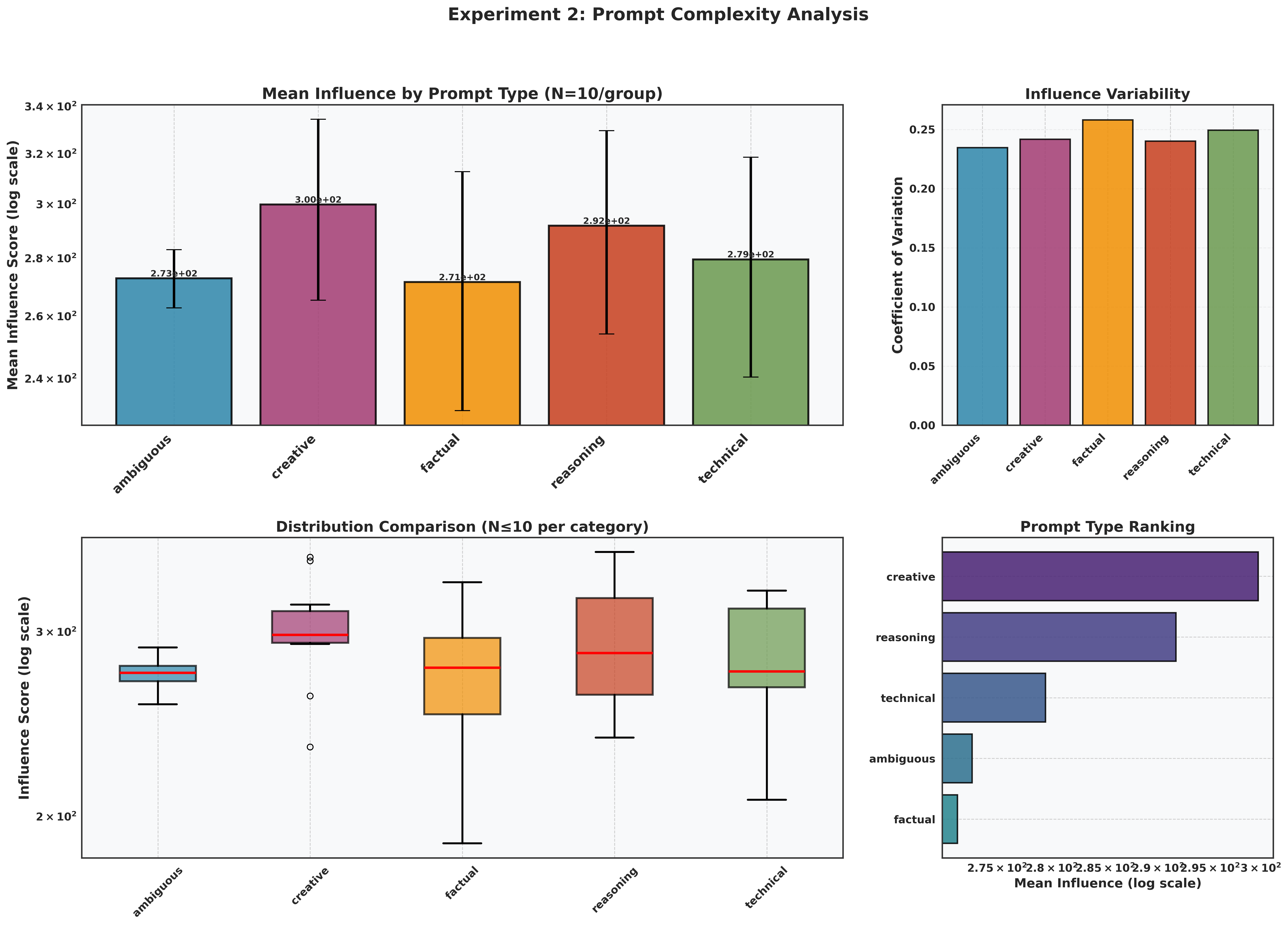}
    \caption{Prompt complexity analysis: influence distribution across prompt types.}
    \label{fig:prompt_complexity_analysis}
\end{figure}

\subsection{Token Type Sensitivity}
\label{subsec:token_type}
\begin{figure}[ht]
    \centering
    \includegraphics[width=0.9\linewidth]{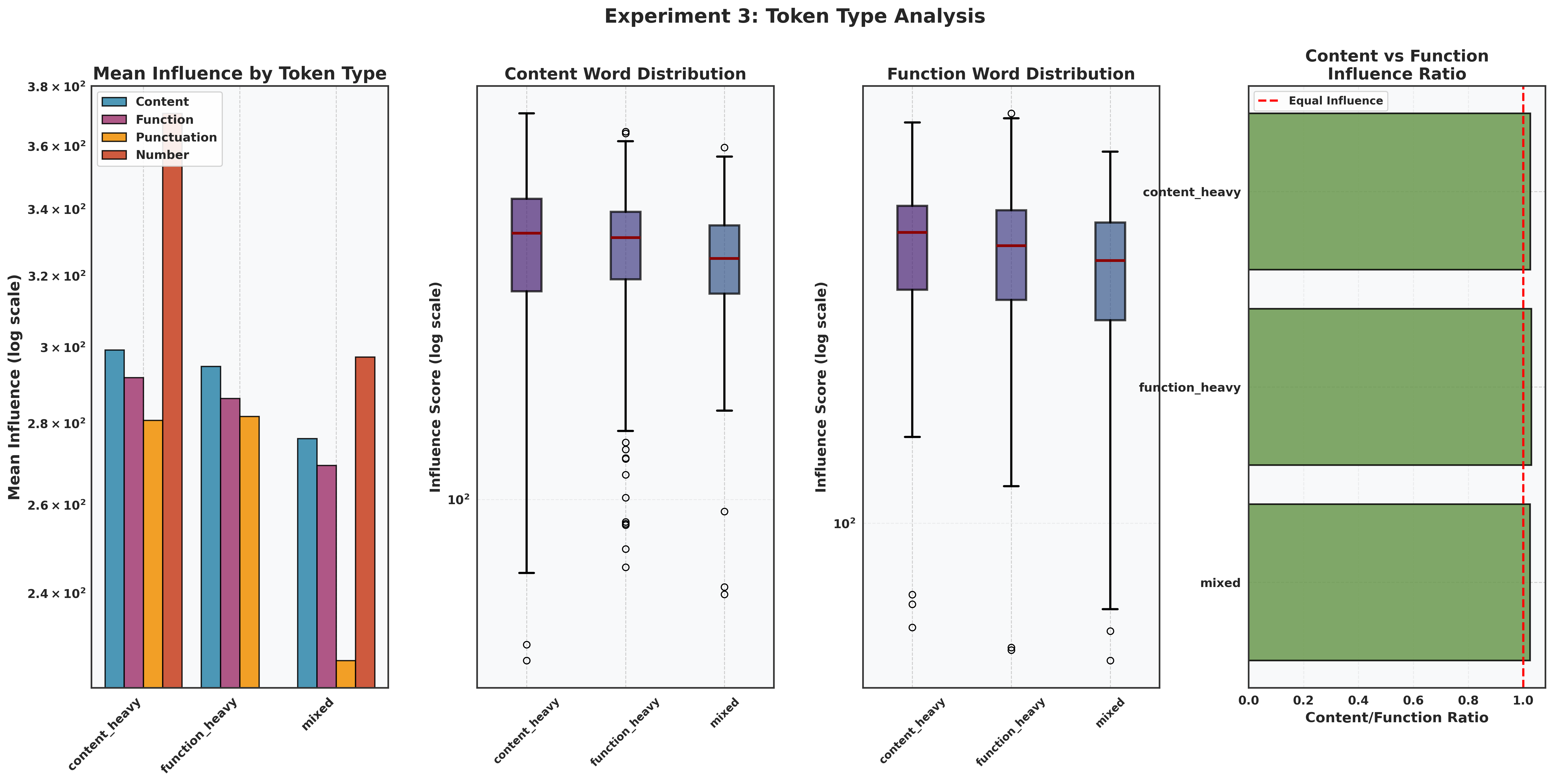}
    \caption{Token-type influence comparison across models. Content tokens maintain consistently higher controllability.}
    \label{fig:token_type}
\end{figure}

Token-level patterns revealed that content tokens consistently exhibited higher influence than function or punctuation tokens across all models (Figure~\ref{fig:token_type}). 
This effect was amplified with scale, with Mamba-2.8B-SlimPJ showing a 15–20\% higher mean influence for content tokens compared to function tokens. 
These findings suggest that the Influence Score is semantically sensitive and emphasizes meaning-bearing tokens.

\subsection{Layer-wise Evolution}
\label{subsec:layer_evolution}
\begin{figure}[ht]
    \centering
    \includegraphics[width=0.9\linewidth]{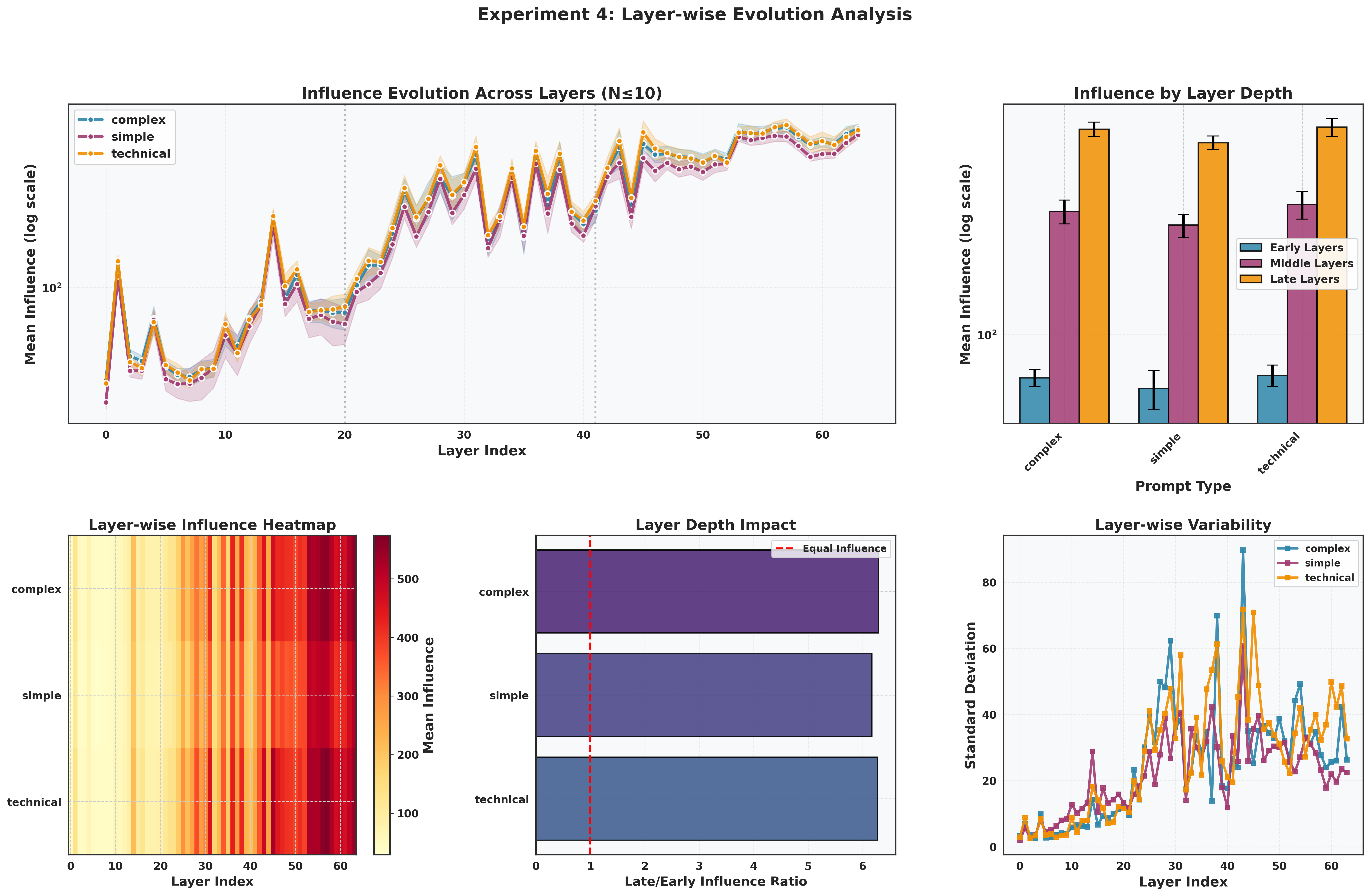}
    \caption{Layer-wise evolution of Influence Score for representative prompts for Mamba-2.8B-SlimPJ model.}
    \label{fig:layer_evolution2.8_s}
\end{figure}

\begin{figure}[ht]
    \centering
    \includegraphics[width=0.9\linewidth]{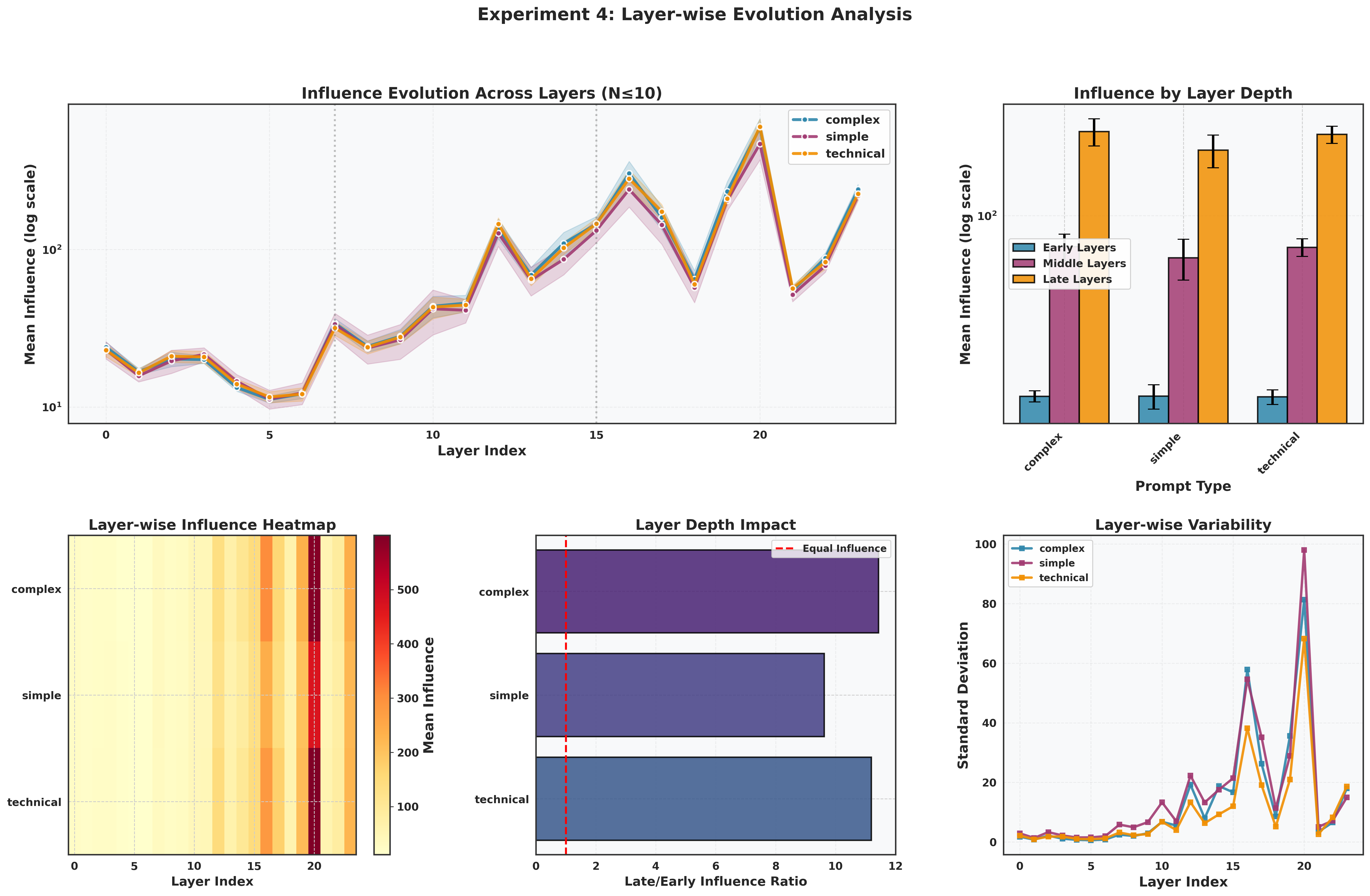}
    \caption{Layer-wise evolution of Influence Score for representative prompts for Mamba-130M model.}
    \label{fig:layer_evolution130}
\end{figure}

A clear monotonic growth in influence was observed across layer depth (Figure~\ref{fig:layer_evolution2.8_s} and Figure~\ref{fig:layer_evolution130}). 
In all models, late layers exhibited 5–11$\times$ higher influence compared to early layers (Table~\ref{tab:layer_evolution}), indicating that Mamba accumulates representational significance progressively across the recurrent SSM blocks. 
This layered buildup supports the hypothesis that the SSM architecture encodes long-range dependencies through gradual internal integration rather than discrete attention jumps.

\begin{table}[!ht]
\centering
\caption{Mean Influence per layer region and Late/Early ratio.}
\label{tab:layer_evolution}
\begin{tabular}{lcccc}
\toprule
\textbf{Model} & \textbf{Early Mean} & \textbf{Mid Mean} & \textbf{Late Mean} & \textbf{Late/Early} \\
\midrule
Mamba-130M & 1.9e+01 & 7.5e+01 & 2.17e+02 & 11.4x \\
Mamba-2.8B & 4.3e+01 & 1.9e+02 & 2.83e+02 & 6.7x \\
Mamba-2.8B-SlimPJ & 7.3e+01 & 2.5e+02 & 4.57e+02 & 6.3x \\
\bottomrule
\end{tabular}
\end{table}

\subsection{Position Sensitivity}
\label{subsec:position_sensitivity}
\begin{figure}[!ht]
    \centering
    \includegraphics[width=0.9\linewidth]{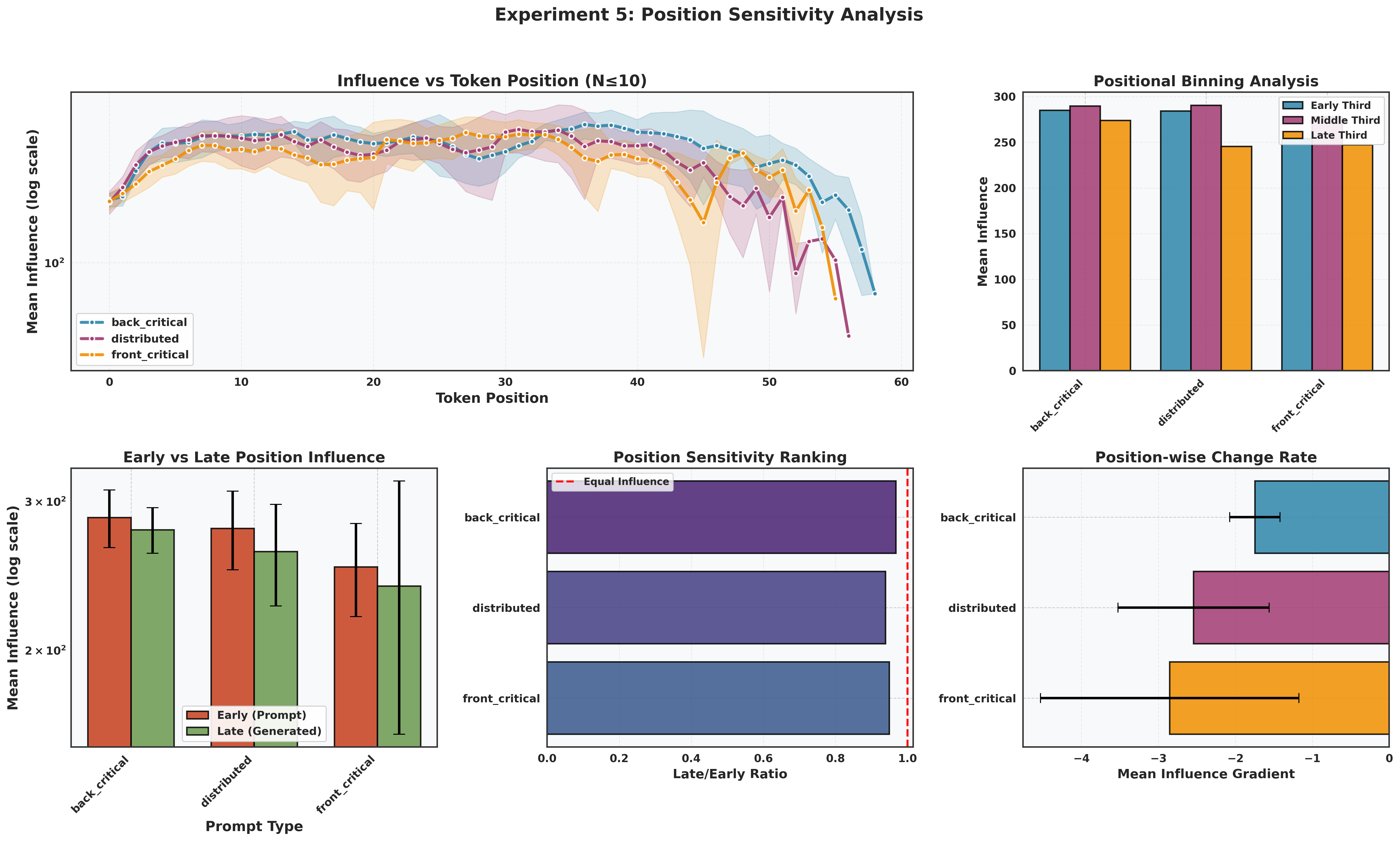}
    \caption{Position sensitivity: average Influence Score for prompt (early) vs. generated (late) tokens.}
    \label{fig:position_sensitivity}
\end{figure}

As shown in Figure~\ref{fig:position_sensitivity}, early (prompt) tokens consistently exerted higher influence than later (generated) tokens. The average late/early ratio remained below 1.0 for all models (0.75–0.97), indicating that SSM models anchor contextual control primarily in the input sequence rather than output autoregression. This stable positional structure contrasts with transformer-based models, which often show mid-sequence peaks.

\subsection{Perturbation Robustness}
\label{subsec:perturbation}
Perturbation tests quantified how textual noise affects controllability. The small Mamba-130M exhibited mild sensitivity (+6.2\% under typos), while the largest model, Mamba-2.8B-SlimPJ, demonstrated strong robustness (-2-7\% change across perturbations). The inverse correlation between model size and perturbation sensitivity implies that larger SSMs develop smoother latent manifolds and stronger compositional stability.

\begin{figure}[!ht]
    \centering
    \includegraphics[width=0.9\linewidth]{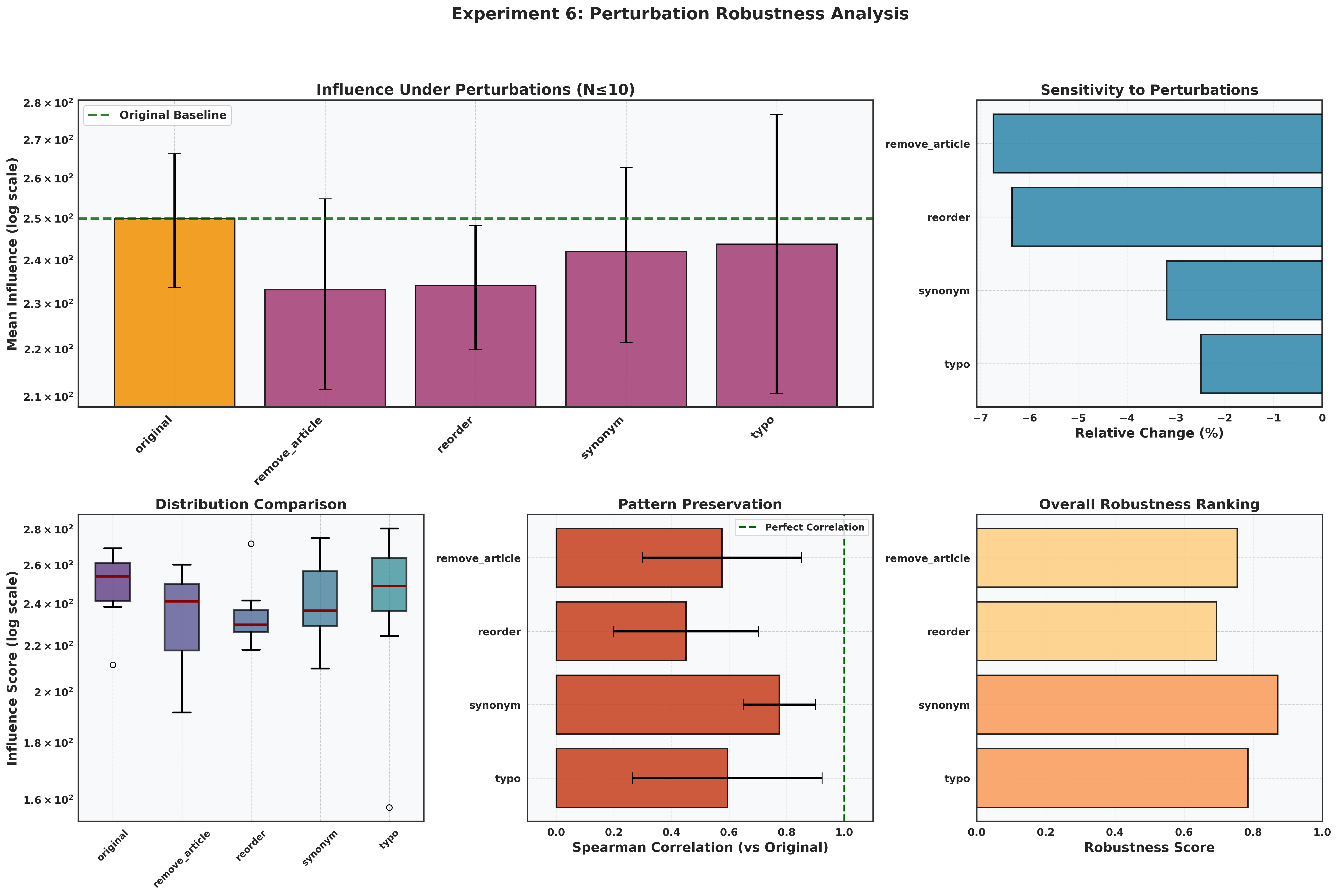}
    \caption{Perturbation robustness across model scales. Bars indicate mean Influence Score deviations under textual perturbations.}
    \label{fig:perturbation}
\end{figure}

\subsection{Cross-Model Scaling Trends}
\label{subsec:scaling_trends}
\begin{figure}[!ht]
    \centering
    \includegraphics[width=0.9\linewidth]{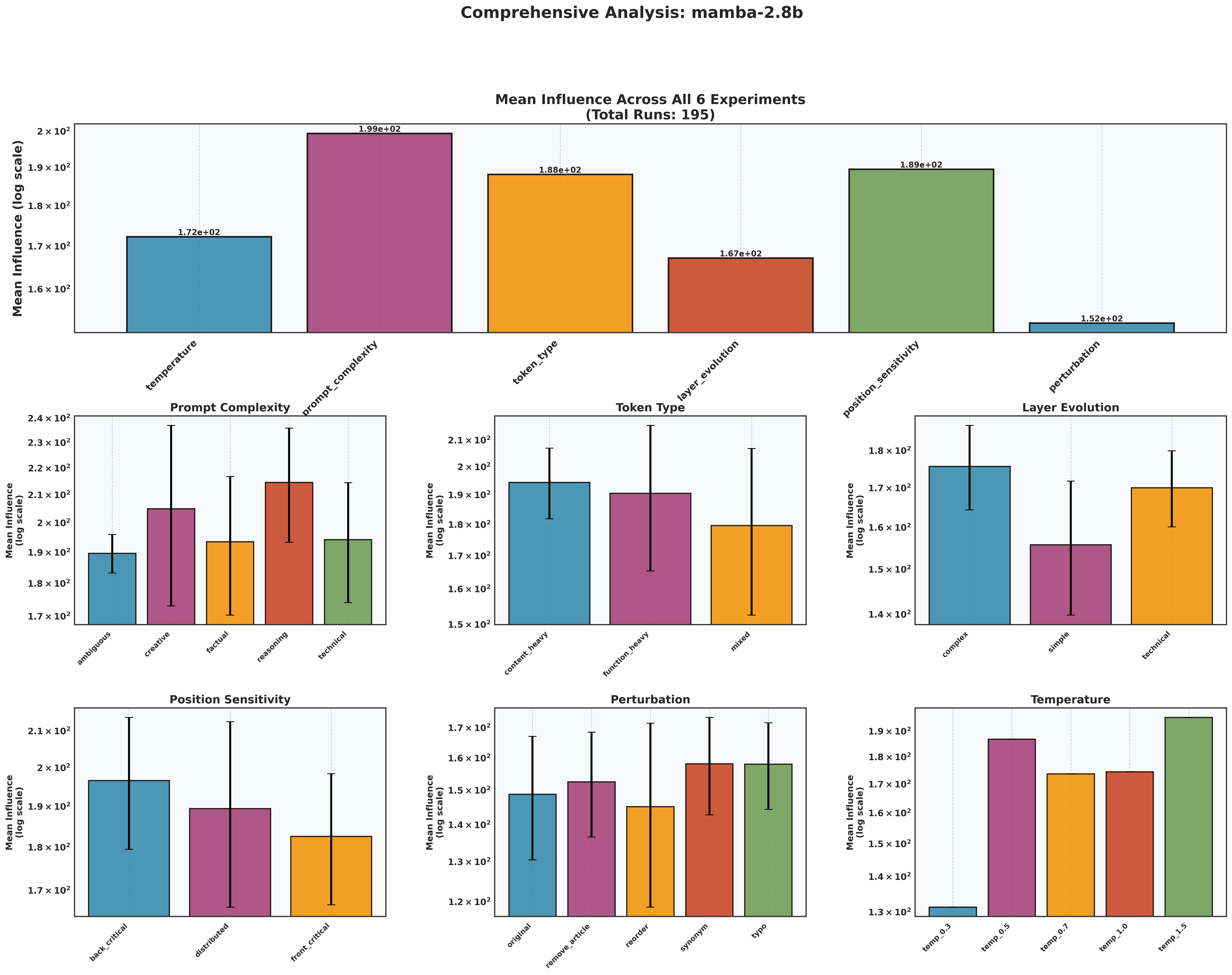}
    \caption{Comprehensive Influence Score comparison across models and experimental categories.}
    \label{fig:comprehensive_comparison}
\end{figure}

Figure~\ref{fig:comprehensive_comparison} summarizes scaling effects across all experiments. The Influence Score exhibits a near-linear growth with parameter count:
\[
I_{\text{mean}} \propto \log(N_{\text{params}})
\]
This scaling law highlights that the representational connectivity measured by the Influence Score increases with model capacity, reinforcing its potential as a general interpretability metric for SSM-based architectures.

\section{Discussion}
\label{sec:disscution}
Our primary objective was to validate the recently introduced \textit{Influence Score}, a metric rooted in control theory, as a diagnostic tool for understanding the internal dynamics of Mamba SSMs. The preceding analysis confirmed its utility, revealing a clear, quantifiable link between model scale, training data, and internal state-space behavior. Our discussion now synthesizes these six experiments into a cohesive interpretation of Mamba's architectural properties and their emergent behaviors.

\subsection{The Influence Score as a Proxy for Model Capacity and Stability}
Our first and most direct finding is that the mean Influence Score scales directly with model size and, crucially, training data quality. The consistent ordering of \texttt{mamba-130m} < \texttt{mamba-2.8b} < \texttt{mamba-2.8b-slimpj} across all experiments is non-trivial. It implies that larger models do not merely possess more parameters; their internal state dynamics are quantifiably more potent. From a control-theoretic perspective, the inputs to the \texttt{slimpj} model have a significantly stronger "control authority" over the system's future state.

This interpretation is powerfully reinforced by the Perturbation Robustness experiment (Exp. 6). The two smaller models exhibited a confusion response: faced with noisy input (e.g., typos), their internal influence increased. This suggests a brittle, unstable system where a disturbance causes a spike in internal effort, likely as the model struggles to reconcile the noise with its learned patterns.

The \texttt{mamba-2.8b-slimpj} model demonstrated the exact opposite, a hallmark of a robust and stable control system: it rejected the disturbance. The decrease in influence for all perturbations suggests the model correctly identified the input as noise and efficiently dismissed it, expending less internal effort. This emergent property of robustnes transitioning from brittle confusion to efficient dismissal is a key qualitative leap captured by our metric.

\subsection{Uncovering Core Architectural Signatures of Mamba}

Beyond behaviors that emerge with scale, the framework reveals core architectural signatures, fundamental properties of the Mamba architecture that persist across all models, from \texttt{mamba-130m} to \texttt{mamba-2.8b-slimpj}. These signatures provide a baseline for understanding how Mamba processes information.

The Layer-wise Evolution, (Exp. 4), analysis provides the clearest evidence of functional specialization. We find that influence is not uniformly distributed across the model's depth; rather, it grows in an exponential-like curve, with influence heavily concentrated in the mid-to-late layers. This pattern is quantified by the Late/Early Ratio (the ratio of mean influence in the final third of layers to the first third). This effect is most pronounced in the 24-layer \texttt{mamba-130m} model, which shows a dramatic 11.43$\times$ increase in influence on complex tasks. The larger 64-layer models stabilize this ratio, but the principle holds: \texttt{mamba-2.8b} shows a $\approx$6.67$\times$ increase and \texttt{mamba-2.8b-slimpj} a $\approx$6.29$\times$ increase. This strongly suggests that, similar to Transformers, Mamba's layers are functionally specialized. The early layers, exhibiting very low influence, appear to function as simple encoders. The cognitively demanding work of processing, refining, and acting upon the model's state is reserved for the mid-to-late layers, which display significantly higher control influence. The metric effectively visualizes this hierarchical thought process.

The Position Sensitivity, (Exp. 5), analysis confirms a core theoretical property of any state-space model. Unlike a global attention mechanism that can be (in principle) permutation-invariant, Mamba's recurrent state mechanism is inherently biased toward recent information. Across all three models, prompts where the critical information was at the end (back\_critical) consistently yielded the highest mean influence scores. Conversely, prompts where the information was at the beginning (front\_critical) yielded the lowest scores. For instance, in the \texttt{mamba-2.8b-slimpj} model, back\_critical prompts had a mean score of 2.82e+02, while front\_critical prompts scored only 2.54e+02. 

This confirms that the control authority of a token is not uniform across the sequence; tokens at the end of the context window exert the most influence on the next-token generation. This has direct, practical implications for prompt engineering: for Mamba-based models, critical instructions should be placed at the end of the prompt for maximum effect. This experiment also confirmed a logical processing hierarchy. For all models, the mean influence of the prompt tokens was significantly higher than the influence of the Generated tokens (Late/Early Ratios were 0.75x to 0.97x). This is an intuitive result: the model correctly identifies the user's prompt as the primary "driver" of the output, rather than its own generated text.

The Prompt Complexity (Exp. 2) analysis reveals a final core signature: Mamba's influence mechanism is a stable generalist. One might hypothesize that a reasoning task would require vastly more internal effort (and thus higher influence) than a simple factual one. The influence scores are remarkably stable across all five task categories. The ratio between the highest-influence task (e.g., creative for \texttt{slimpj}) and the lowest-influence task (e.g., factual for \texttt{slimpj}) is extremely tight: \textbf{1.11$\times$} for \texttt{mamba-2.8b-slimpj} and \textbf{1.13$\times$} for \texttt{mamba-2.8b}. This demonstrates that the core influence mechanism is not a brittle specialist that over-reacts to specific task types. Instead, it acts as a stable generalist, applying a consistent level of internal processing regardless of the linguistic or cognitive domain. This baseline stability is a fundamental characteristic of the architecture.

\subsection{Limitations and Future Directions}
While this work validates the Influence Score, it also opens new lines of inquiry. The score is an aggregated metric; future work should trace the influence of individual tokens or concepts through the network on a layer-by-layer basis. Furthermore, this study was limited to three models; applying this framework to a wider range of SSMs and comparing their influence fingerprints to traditional Transformers would be a fruitful endeavor.

A significant future direction lies in guided generation. If, as our data suggests, pathological outputs are preceded by anomalous influence scores (e.g., the confusion spikes in Exp. 6), it may be possible to steer the generation process. One could penalize decoding paths that lead to a sudden, unstable spike in system influence, thereby improving model reliability by enforcing a more stable, controllable internal state.

\bibliographystyle{unsrt}  
\bibliography{references}  
\appendix


\end{document}